\pdfoutput=1
\PassOptionsToPackage{table}{xcolor} 
\documentclass[11pt]{article}
\usepackage{EMNLP2023}
\usepackage{times}
\usepackage{latexsym}
\usepackage[T1]{fontenc}
\usepackage[utf8]{inputenc}
\usepackage{microtype}
\usepackage{multirow, makecell}
\usepackage{inconsolata}
\usepackage{tikz}
\usepackage{amsmath}
\usepackage{soul}
\usepackage{subfig}
\usepackage{hyperref}
\usepackage{booktabs,tabularx}
\usepackage{censor}

\title{Tracking the Newsworthiness of Public Documents}

\author{Alexander Spangher$^a$\thanks{Corresponding Author: \texttt{spangher@usc.edu}}, Emilio Ferrara$^a$, Ben Welsh$^b$, \\ \bf Nanyun Peng$^c$, Serdar Tumgoren$^d$, Jonathan May$^a$ \\
  $^a$ Information Sciences Institute, University of Southern California \\
  $^b$ Reuters News, $^c$ University of California, Los Angeles, $^d$ Stanford University \\
  \\}

\begin{document}
\maketitle
\begin{abstract}

Journalists must find stories in huge amounts of textual data (e.g. leaks, bills, press releases) as part of their jobs: determining \textit{when} and \textit{why} text  becomes news can help us understand coverage patterns and help us build assistive tools. Yet, this is challenging because very few labelled links exist, language use between corpora is very different, and text may be covered for a variety of reasons.
In this work we focus on news coverage of local public policy in the San Francisco Bay Area by the \textit{San Francisco Chronicle}. 
First, we gather news articles, public policy documents and meeting recordings and link them using \textit{probabilistic relational modeling}, which we show is a low-annotation linking methodology that outperforms other retrieval-based baselines. 
Second, we define a new task: \textbf{\textit{newsworthiness prediction}}, to predict if a policy item will get covered. We show that different aspects of public policy discussion yield different newsworthiness signals. 
Finally we perform human evaluation with expert journalists and show our systems identify policies they consider newsworthy with 68\% F1 and our coverage recommendations are helpful with an 84\% win-rate.

\end{abstract}

\vspace{.3cm}
\section{Introduction}

Press coverage of local policy can be crucial for the health of a community: it can increase civic engagement \cite{smith1987effects}, improve government \cite{bruns2011newspaper} and engender greater work output from politicians \cite{snyder2010press}. 
Yet sparse quantitative work exists analyzing such coverage: (1) \textit{what} policy gets covered, (2) \textit{why} it got covered, and (3) what \textit{impact} does the coverage has?
Determining that a specific policy item\footnote{i.e. A motion of gov.: a proposal, bill, 
settlement, 
etc..} was covered in media, as shown in Figure \ref{fig:example}, is a challenging task. Unlike related tasks, like \textit{citation prediction} \cite{shibata2012link} or \textit{cross document event-coreference} \cite{bagga1999cross}, determining policy coverage requires us to establish links between documents in two different linguistic domains, with no pre-existing labels.

\begin{figure*}[t]
\includegraphics[width=\linewidth]{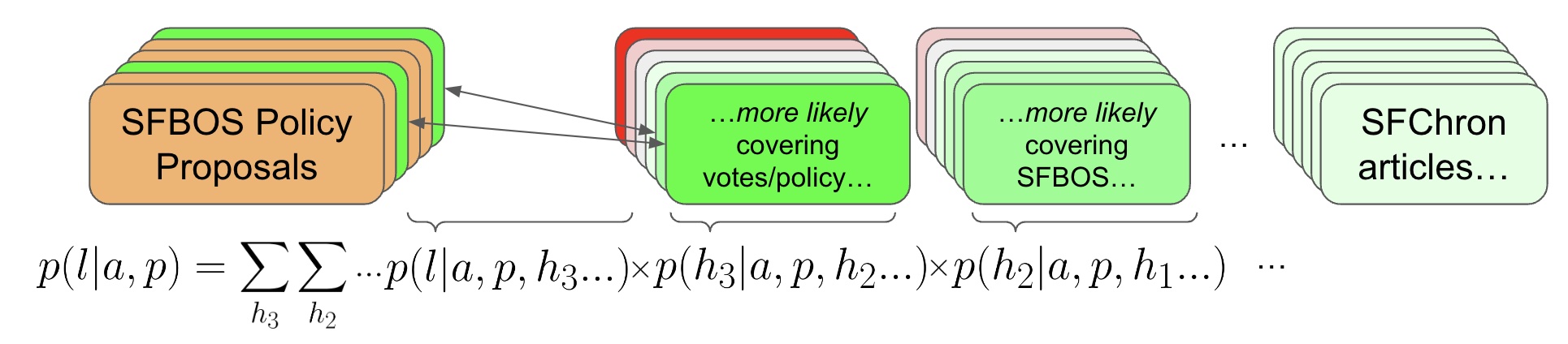}
\caption{Our probabilistic relational modeling (PRM) process for whether an article $a$ covers a city council proposal, $p$, i.e. are linked, $l$. PRM works by introducing auxiliary marginal variables $h_1, ...h_n$ that refine the link model, $p(l | a,p)$ through conditioning. In the diagram, moving from right-to-left, each step shows another variable $h_i$ being applied in the PRM-chain: e.g. $h_2=$\textit{``covering SFBOS''}, $h_3=$\textit{``covering SFBOS votes and policy''}. $h_2, h_3$, etc. can be learned separately, and we learn supervised models for each step.}
\label{fig:matching-process}
\end{figure*}

In this work, we build methods to establish when a news article covers a local policy document, i.e. to \textit{link} them. We show that breaking this problem down into a chain of decisions, each conditional on the previous\footnote{Shown in Figure \ref{fig:matching-process}, i.e. ``article covers local politics'' $\rightarrow$ ``article covers city council meetings'' $\rightarrow$ ``covers past meeting'' $\rightarrow$ ``covers \textit{this} past meeting''}, an application of probabilistic relational modeling (PRM) \cite{getoor2002learning}, helps us outperform other retrieval-based baselines. 

\begin{figure}
\begin{tikzpicture}
  \node [draw, text width=7cm, align=left, fill=blue!20, rounded corners] (box1) at (0,0) {Mandelman Ordinance amending the Planning Code to increase density on lots with auto-oriented uses...};

  \node [above, anchor=west, yshift=6pt] at (box1.north west) {\large \textbf{Policy Document}};

  \node [draw, text width=7cm, align=left, fill=yellow!30, rounded corners] (box2) at (0,-3.1) {After 14 months of delays, the Board of Supervisors on Tuesday unanimously passed Mayor Breed’s legislation that makes it easier to turn gas stations, parking lots and other auto-related properties into housing. This caused widespread debate....};

  \node [above, anchor=west, yshift=6pt] at (box2.north west) {\large \textbf{News Article}};

  \draw [<-, line width=2.5pt] (box1) -- (box2);
\end{tikzpicture}
\caption{In this paper, we establish the \textit{newsworthiness prediction} task. We (1) train models to infer when public policy items have been covered in the press and (2)  predict if new items \textit{will} be covered. }
\label{fig:example}
\end{figure}

Next, having established links, we seek to predict what \textit{new} policy will get covered. More specifically, we propose a new task, \textit{newsworthiness prediction}, to predict whether a novel policy proposal should get covered, \textit{\ul{according to previous coverage patterns}}. Addressing this task is useful: experiments with human journalists show us that our newsworthiness models surface relevant items.

Finally, we ask whether policy text alone is enough to predict coverage. We study recordings of public meetings where policy proposals are addressed. We find that policy items that get covered in news media also get discussed longer in meetings and have more members of the public addressing them during public comment periods. 

We start small in order to prove that this task is viable, focusing on one locality, San Francisco Board of Supervisors (SFBOS) and it's coverage in the the San Francisco Chronicle (SFChron). This means that our analysis \textit{\ul{is necessarily limited to ``newsworthiness'' as defined by the SFChron}}, on SFBOS policy text. However, the recipe that we explore here in one locality is largely applicable to any locality or topic domain and we intend in future work to expand the corpora we consider. 

In sum our contributions are:
\begin{enumerate}
    \item We collect a large multimodal dataset of 13,000 SFBOS policy proposals spanning 10 years, 20,000 SFChron news articles and 3,400 hours of SFBOS meeting video (which we transcribe and diarize), in Sections \ref{sct:sct-1:dataset}, \ref{sct:sct-2:dataset} in order to study newsworthiness \textit{\underline{in the local context of one city}}.
    
    \item We link these corpora with a novel application of probabilistic relational modeling, outperforming modern baselines (Section \ref{sct:sct-1:sublinks}). We find that between 2-6\% of SFBOS policies get covered in SFChron (Section \ref{sct:sct-1:analysis}).
    
    \item We establish a novel task, \textit{newsworthiness prediction}, and use it to analyze what makes policy and public discussion newsworthy, finding that newsworthiness is predictable and has a strong non-temporal element (Section \ref{sct:sct-2:results}). We show journalists find our rankings helpful in surfacing newsworthy leads.
\end{enumerate}

\section{Policy Item $\leftrightarrow$ Article Linking}
\label{sct:sct-1}

Our goal in this section is to determine which articles cover which policy items. We seek to model the likelihood a link $l$ exists between an article, $a$, and a specific policy item, $p$, or $P( l | a, p )$.

We apply the \textit{probabilistic relational model} (PRM) framework \cite{getoor2002learning} to solve this problem. In PRM, we learn conditional attributes  $h_1, ... h_t$ of one or both items and marginalize, or:
\begin{equation}
\begin{split}
    P( l | a, p ) = \\ 
    \sum_{h_1} \ldots \sum_{h_t} p(l | a, p, h_1, \ldots,  h_t) \ldots p( h_1 | a, p )
\label{eq:factorized-link}
\end{split}
\end{equation}
Where, as shown in Figure \ref{fig:matching-process}, $h_2$ might be ``covers SFBOS'', and $h_3$ might be ``covers SFBOS votes/policy.''\footnote{Because no natural linking information exists (i.e. hyperlinks in the article body), we typically model $l_*$ on the text of the article and/or policy proposal.} Not all politics articles are about SFBOS, and not all SFBOS articles cover policy. Such variety confounds unsupervised models, but is solvable when broken into easier-to-supervise subproblems. This is not dissimilar to Chain-of-Thought (CoT) \cite{wei2022chain}, where language models decompose complex reasoning tasks. 
%
%
%

\subsection{Corpora: SFBOS Policy-Proposals and SFChron Articles}
\label{sct:sct-1:dataset}

We focus on a specific local government, SFBOS, and a specific newspaper, SFChron, that has a robust local news section. We start by gathering HTML of all SFChron articles published between 2013--2023 and via the Common Crawl\footnote{We search for all URLs matching wildcard pattern \texttt{https://www.sfchronicle.com/*}}. We parse article text\footnote{Using the \texttt{Newspaper} package: \url{https://github.com/codelucas/newspaper}.} and deduplicate based on text, and ultimately are left with a set of 202,644 SFChron articles\footnote{We release the full list of URLs in our experiment, as well as scripts to replicate our collection process.}. We also scrape the public meeting calendar on the SFBOS website\footnote{\url{https://sfgov.legistar.com/Calendar.aspx}} to collect all SFBOS meetings between 2013-2023\footnote{Example meeting: \url{https://sfgov.legistar.com/MeetingDetail.aspx?ID=1108038&GUID=8B3A2668-90A9-43E9-A694-8747176617F4}} and then collect the proposal text for 13,089 SFBOS policy proposals\footnote{Example of a policy proposal: \url{https://sfgov.legistar.com/LegislationDetail.aspx?ID=6251774&GUID=420031B2-94DE-440F-AB74-25FF091F2D61}} that were discussed a total of 27,371 times in 410 public meetings. Each policy is, on average, discussed in 3 separate SFBOS meetings.

\subsection{Devising Relational Chains}
\label{sct:sct-1:sublinks}
We manually identify steps shown in Figure \ref{fig:matching-process} and hand-craft models for each attribute. 
We hire two journalists\footnote{We adjust pay to be roughly \$20 USD an hour.} to annotate data for each $h_i$. 
\begin{enumerate}
    \item $h_1$: ``$a$ covers SFBOS''. We use the keyword $t = $``Board of Supervisors'' to identify candidates. Then we delete $t$ from these candidates, sample negatives and bootstrap a classifier to identify more candidates. Our annotators label 100 of these and we train a classifier $p(h_1 | a)$.
    \item $h_2$: ``$a$ covers votes/policy''. From $h_1$ articles, our journalists label an additional 100 articles on whether they mention votes and policy. We train a classifier $p(h_2 | h_1, a)$.
    \item $h_3$: ``$a$ covers recent policy from SFBOS''. We use GPT3.5 with a 10-shot prompt to determine whether $a$ mentions votes occurring less then a month prior to publication. We also ask GPT3.5 to confirm the government body is SFBOS (e.g. not ``Oakland City Council''). We use logits for ``yes''/``no'' as $p(h_3 | h_2, h_1, a)$.
    \item $l$: ``$a$ covers policy $p$.'' We match articles to city-council meeting minutes using cosine similarity over the vector space.
\end{enumerate}

Our link prediction problem is many-to-many, so we wish to choose a threshold, $\lambda$, for Equation \ref{eq:factorized-link}, to consider items a match. To help us choose $\lambda$ and to evaluate our method, our annotators manually identify 100 true pairs, which we split 50/50 into $S_{gold, train}$ and $S_{gold, test}$.

\begin{table*}
    \centering
    \begin{tabularx}{\linewidth}{Xrrr}
    \toprule
    PRM-Chain & TF-IDF  & SBERT & OpenAI Embeddings\\
    \midrule
    $p(l | a, p)$, base & 16.0 & 32.1 & 30.3 \\ 
    $\sum_{h_1} p(l | a, p, h_1) p(h_1 | a, p)$ & 28.5 & 33.9  & 37.5 \\ 
    $\sum_{h_1, h_2} p(l | a, p, h_1, h_2) p(h_2 | h_1, a, p)...$ & 55.3 & 48.2 & 53.5 \\
    $\sum_{h_1, h_2, h_3} p(l | a, p, h_1, h_2, h_3) p(h_3 | h_1, h_2, a, p)...$ & \textbf{68.2} & 55.6 & 62.6 \\
    \bottomrule
    \end{tabularx}
    \caption{Results from training PRM chains over different sentence embeddings. $l$ is defined as a mapping between News article $a$ $\leftrightarrow$ Policy mapping $p$. We establish a score-threshold for $p(l | a, p)$ for each trial using our gold-labeled dataset, $S_{gold, train}$ and report f1-scores using $S_{gold, test}$. TF-IDF is defined \newcite{ramos2003using}. SBERT uses the \texttt{all-MiniLM-L6-v2} model \cite{reimers2019sentence}. OpenAI uses the \texttt{text-embedding-ada-002} model.}
    \label{tbl:part-1-results}
\end{table*}

\subsection{Linking Results}
\label{sct:sct-1:results}

Our attribute-based model, as shown in Table \ref{tbl:part-1-results}, helps us retrieve $(a, p) \in S_{gold}$ with 68\% f1. We show via an ablation experiment that each attribute $h_i$ is important for our final prediction: Table \ref{tbl:part-1-results} shows how F1 drops from 68\% to 16\% when we remove $h_i$-conditioning steps.

Surprisingly, using PRM with TF-IDF outperforms different embedding methods like SBERT \cite{reimers2019sentence} and OpenAI embeddings \cite{openaiembeddings}. We suspect that specific technical phrases are important for this task, which unsupervised embeddings might ignore; training a supervised retrieval architecture like DPR might help represent these phrases in the embeddings, but requires 100-1000 times more data than we have collected \cite{karpukhin2020dense}. 
We note an attribute-based linking approach also lends well to CoT \cite{wei2022chain}. As language models become cheaper and more scalable, this could be a viable end-to-end approach.

Despite our positive results, we acknowledge that \textit{\ul{our approach is limited}}. If there is no lexical overlap between $a$ and $p$, we would not discover a link even if there were one. Also, we might be more exposed to this risk than the results show: in constructing $S_{gold}$, our annotators might have also faced a similar bias depending on the retrieval mechanisms (e.g. search) they used. \textit{\ul{A more comprehensive evaluation set would be generated by journalists \textit{as they are working} on stories.}} We discuss further limitations in Section \ref{sct:discussion-and-conclusion}.

\subsection{Linking Analysis}
\label{sct:sct-1:analysis}
We scale our models across our entire corpus of SFChron and SFBOS articles from 2013-2023. Examining these links gives us insights into the amount of coverage devoted to public policy.

\paragraph{Roughly 7.8\% of SFBOS policy proposals get covered,} or 1,105 out of 13,089 policies. These policies are covered by a total of 1,828 news articles. Although each policy is covered on average 1.8 times, the distribution is right-skewed and the median coverage is one time per policy. See Appendix \ref{app:descriptive-statistics}, Figure \ref{fig:news-articles-per-policy-item} for more details. The policies that are covered many times are a mixture of staffing (e.g. ``Nomination of a Successor Mayor''), transportation bills (e.g. ``Unauthorized scooter violations'') and emergency ordinances (e.g. ``COVID-19 Safe Shelter Operations''.) Again, see Appendix \ref{app:descriptive-statistics}, Table \ref{tab:top-covered-bills}.

\begin{figure}[t]
    \centering
    \includegraphics[width=\linewidth]{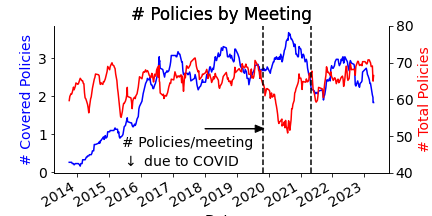}
    \caption{The number of policies introduced to SFBOS and those covered by SFChron, measured by the date the policy was introduced and whether $p(a,p)>\lambda$.}
    \label{fig:num-articles-over-time}
\end{figure}

\paragraph{Coverage of policies is constant across time.}
As shown in Figure \ref{fig:num-articles-over-time}, between 1--3 policies are covered per meeting, out of between 50--60 presented. This equates to between 2\%--6\% of proposals being covered consistently throughout our 10 years window. Coverage is relatively constant throughout the observation period, removing newspaper decline\cite{mathews2022life} as a possible confounder to newsworthiness decisions\footnote{Due to an ongoing economic crisis in journalism, many newspapers are shrinking, leading to less coverage and, possibly, changes in what is considered ``newsworthy''.}.
We observe a brief spike in \% of $p$ covered in 2020--2021. Closer examination reveals that the number of $(a, p)$ pairs stay constant, however, the number of $p$ \textit{proposed} drops significantly, from an average of 64 policies per meeting prior to 2021, to 52 policies per meeting in the first 9 months of 2021. This is likely the result of COVID-related shutdowns. Conversations with SFChron journalists confirm this. 

\section{Newsworthiness Prediction}
\label{sct:sct-2}

Next, we wish to ask \textit{why} certain policy proposals are covered. To address this, we establish a new task, \textit{newsworthiness prediction}: predict, given a policy item $p$, if an article will write about it. We use our linked dataset  $\{(a, p)\}$, in Section \ref{sct:sct-1}, and treat this problem as a prediction problem where:
\begin{equation}
Y(p) =
\begin{cases}
  1, & \text{ if } p \in \{(a, p)\} \\
  0, & \text{otherwise}
\end{cases}
\label{eq:newsworthiness-label}
\end{equation}

Our goal is twofold: (1) Learning a good model can show us which features of policy-items lead to coverage. (2) Performing this task well at inference time takes us steps closer to building tools that will be useful for surfacing potential stories.

Previously, \textit{newsworthiness} has been addressed as a feature-detection problem, as in \cite{diakopoulos2010diamonds}, where engineered-features measured specific criteria\footnote{E.g. ``statistically anomalous'' \cite{zhao2014fluxflow}, ``sentiment=happy'')}. Journalists examined combinations of features to find newsworthy items but could miss items if their newsworthiness did not fit the measurements. Because we formulate our task as a prediction task, backed by a dataset, we can also expose new and possibly unexpected features. \textit{\ul{However, a prediction-based approach is limited in its own ways.}} We assume that past coverage patterns predict future patterns, and that journalists generally agree. We will explore these assumptions in Section \ref{sct:sct-2:results}, and we will also see notable cases where these assumptions \textit{do} limit us.

\begin{table}[t]
    \centering
    \begin{tabularx}{\linewidth}{>{\hangindent=2em}X}
        \toprule
        Policy Features Analyzed \\
        \midrule
        text of proposal \\
        \# prior meetings proposal has been discussed\\
        \# prior news articles linked to proposal\\
        \midrule
        length of time proposal is discussed in meeting \\
        transcribed text of city-council member's policy discussion\\
        \# public commenters discussing the policy\\
        summary of public commentary\\
        \bottomrule
    \end{tabularx}
    \caption{Summary of features for each policy item. Top section is generated via $(a, p)$. Bottom section is generated via SFBOS video transcriptions.}
    \label{tab:features}
\end{table}

\subsection{Newsworthiness Training Corpus}
\label{sct:sct-2:dataset}

We extract features from the linked $(a, p)$ pairs derived in the first section to construct our training corpus. As shown in Figure \ref{fig:example}, in the news article, there are remarks: ``After 14 months of delay'', ``widespread debate'' that seem to indicate that there aspects of this policy that are \textit{not} solely related to its topic that made it newsworthy.

To capture some of these features, we include SFBOS meetings where these policies are discussed. We download audio for all meetings in our corpus\footnote{Example: \url{https://sanfrancisco.granicus.com/player/clip/43908}.} and we use the WhisperX package \cite{bain2022whisperx} to transcribe and perform speaker-diarization
. See Appendix \ref{app:aligning-meeting-transcripts} for more about aligning transcripts. We associate each $(a,p)$ with a specific meeting if: (1) $p$ is discussed in the meeting and (2) $a$ was published within a month of the meeting occurring.

Finally, in every SFBOS meeting, there is a special time for members of the public to speak, called ``Public Comment''. Since good newswriting is emotional \cite{uribe2007aresensational}, we hypothesize that ``Public Comment'' might offer an additional lens on a policy's newsworthiness. We determine which speakers are members of the public using diarization to identify speakers that \textit{only} spoke during ``Public Comment''\footnote{We infer the sections of the transcript like ``Public Comment'' using time-stamped agendas, see Appendix \ref{app:aligning-meeting-transcripts} for more detail.}. Then, we calculate the lexical overlap between their speech and the policy text. For more details about ``Public Comment'' and other meeting sections, please see Appendix \ref{app:additional-meeting-eda}. We show all of the features that we use for newsworthiness prediction in Table \ref{tab:features}.




\begin{table*}\
\centering
\begin{tabular}{llllllll}
\toprule
\multicolumn{8}{c}{$\Delta$ Word Distributions for Newsworthy vs. Non-Newsworthy Text} \\
\midrule
\multicolumn{4}{c}{Policy Text} & \multicolumn{2}{c}{Meeting Speech} & \multicolumn{2}{c}{Public Comment} \\ 
\cmidrule(lr){1-4} \cmidrule(lr){5-6} \cmidrule(lr){7-8}
\cmidrule(lr){1-2} \cmidrule(lr){3-4} \cmidrule(lr){5-6} \cmidrule(lr){7-8}
authorizing & \cellcolor{red!41} -0.41 &       housing & \cellcolor{blue!35} 0.35 & supervisor & \cellcolor{green!66} 1.98 &      budget &  \cellcolor{purple!40} 0.4 \\
     county &  \cellcolor{red!30} -0.3 &        health & \cellcolor{blue!31} 0.31 &      think & \cellcolor{green!29} 0.89 & philippines & \cellcolor{purple!16} 0.16 \\
      grant & \cellcolor{red!26} -0.26 &         board &  \cellcolor{blue!30} 0.3 &       know & \cellcolor{green!27} 0.82 &       solar & \cellcolor{purple!15} 0.15 \\
    lawsuit & \cellcolor{red!25} -0.25 &     ordinance & \cellcolor{blue!28} 0.29 &       want & \cellcolor{green!26} 0.78 &     medical & \cellcolor{purple!15} 0.15 \\
      bonds & \cellcolor{red!23} -0.23 &         covid & \cellcolor{blue!28} 0.28 &     people & \cellcolor{green!25} 0.76 &       covid & \cellcolor{purple!14} 0.14 \\
 settlement & \cellcolor{red!22} -0.22 &    department & \cellcolor{blue!23} 0.23 &       like & \cellcolor{green!19} 0.58 &    caltrain & \cellcolor{purple!14} 0.14 \\
   contract & \cellcolor{red!21} -0.21 &      cannabis & \cellcolor{blue!22} 0.22 &       need & \cellcolor{green!14} 0.43 &       rooms & \cellcolor{purple!13} 0.13 \\
     expend & \cellcolor{red!19} -0.19 &      election & \cellcolor{blue!21} 0.21 &  president & \cellcolor{green!12} 0.37 &  amendments & \cellcolor{purple!12} 0.12 \\
\bottomrule
\end{tabular}
\caption{Most likely words associated with newsworthy policy proposals, meeting speech and public comment, measured by $p(w | Y(p)=1) - p(w | Y(p)=0)$, where $p(w | .)$ is based on observed word counts. Also shown in the left-most column is the \textit{least} likely words (negative-valued). Colors shown are a heatmap for easy viewing.}
    \label{tab:most-likely-words}
\end{table*}


\begin{figure}[t]
    \centering
    \includegraphics[width=.9\linewidth]{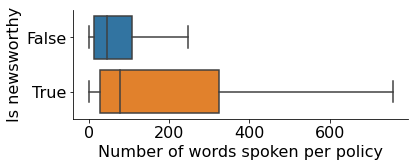}
    \caption{Number of words spoken per meeting for newsworthy policies versus non-newsworthy policies.}
    \label{fig:newsworthiness-time-spent}
\end{figure}

\subsection{Newsworthiness Descriptive Analysis}

Before showing results from the predictive modeling, we show descriptive results. Our main takeaway from this section is that policy text, meeting text and public speakers each are conveying \textit{different} newsworthiness information. 

\paragraph{Policy Text, Meeting Speech and Public Comment all cover different newsworthy topics.}
\label{sct:newsworthiness:textual-analysis}
We see a clear pattern in the kinds of words and topics used in newsworthy policies, meeting speech and public commenters. Table \ref{tab:most-likely-words} shows the top most likely words in each aforementioned text category, calculated as $\Delta p(w) = p(w | Y(p)=1) - p(w | Y(p)=0)$. In the written policy text, we observe topic-specific words like ``housing'', ``covid'' and ``cannabis'' more in newsworthy policies. Topics that were more likely to receive coverage, shown in Table \ref{tbl:top-topics}, include ``Hearings'' and ``Environment''. However, meeting speech for newsworthy policies (which is primarily speech of the SFBOS Supervisors and staff) is directed at deliberation, like ``think'' and ``know''. Finally, during public comment, we see topic-specific speech, but related to a different set of concerns, like ``solar'', ``caltrain'', ``hotels''. We hypothesize that these are each different aspects of newsworthiness that are being conveyed.

\paragraph{Newsworthy Policies are addressed for longer at meetings, by more people.}
Policies that end up getting covered in SFChron are also discussed at greater length than policies that are not: this includes (1) more words spoken (Figure \ref{fig:newsworthiness-time-spent}), (2) more minutes spent discussing (7.7 minutes vs. 2.1), and (3) more speakers spent addressing it (4 speakers vs. 2.2. This number includes members of the public and council members.)
\footnote{Journalists gave us initial feedback, saying that city councils sometimes shove important policies into sections of the meeting like ``Consent Calendar'' and ``Roll Call'', which are typically \textit{not} addressed for a long period of time. This implies either that these cases are truly a minority, or that not enough attention is being paid to these sections of the meeting.
}.  

The number of public commenters we are able to associate with specific policies, on the other hand, is a relatively small number. We are only able to establish an expected $n=.06$ speaker per newsworthy policy and $n=.04$ speaker per non-newsworthy policy. This amounts to 768 speakers associated, overall, with 13,089 policies. Thus, we hypothesize that public comment will not impact our modeling performance, despite observations in Figure \ref{tab:most-likely-words} that public commenters tend to speak to different topics. \textit{\ul{We acknowledge this as yet another limitation of our work and dataset.}} We hope that future work can either (1) establish better methodologies to associate more public commenters with policies (2) collect larger public meeting datasets or (3) incorporate other channels (e.g. social media).

\begin{table}[t]
    \begin{tabularx}{\linewidth}{>{\hangindent=1.2em}X}
        \toprule
        Full Prompt Example\\
        \midrule
        (1) Policy description: "Priority for Veterans with an Affordable Housing Preference under Administrative..." \\
        Presented in 2 prior meetings, 0 news articles \\ 
        \cmidrule(lr{.5\linewidth}){1-1}
        (2) Introduced by 4 speakers in the meeting for 0.7 minutes:\\
        "...Without objection, this ordinance is finally passed unanimously. Madam Clerk..."\\
        \cmidrule(lr{.5\linewidth}){1-1}
        (3) 1 members of the public spoke for 1 minutes.\\
        "<SPEAKER 1> spoke for 1 minutes and said: "Hello, this is \censor{Jonathan Randolph}. I would like to oppose the motions affirming..."\\
        Is this newsworthy? Answer "yes" or "no".\\
        \bottomrule
        \end{tabularx}
    \caption{Example prompt that shows 3 primary components: (1) \textbf{Policy text}, (2) \textbf{Meeting text} and (3)\textbf{ Public commentary text} (name censored). Text is truncated at first 50 words. Further truncated in this example for brevity. Section lines/numbers shown for clarity.}
    \label{tbl:prompt-example}
\end{table}

\begin{table}[t]
    \centering
    \begin{tabular}{lrrrr}
        \toprule
         Model & F1 & ROC & R@10 & MRR \\
         \midrule
         \multicolumn{4}{l}{Fine-tuned GPT3-Babbage} & {}  \\ 
         \midrule
         \quad full & \textbf{25.1} & \textbf{75.9} & \textbf{64.1} & \textbf{29.2} \\
         \cmidrule(lr){1-1}
         \quad (1), (2) & 24.2 & 71.2 & 63.1 & 27.2 \\
         \quad (1)  & 16.2 & 64.5 & 52.2 & 23.1 \\
         \cmidrule(lr){1-1}
         \quad (2), (3) & 14.4 & 57.6 & 37.2 & 15.9 \\
         \toprule
         LR, full & 19.7 & 67.3 & 51.1 & 22.8 \\
         \midrule
         GPT4, full & 18.4 & 62.6 & 40.6 & 16.2 \\
         GPT3.5, full & 13.4 & 63.2 & 46.7 & 21.3 \\
         \bottomrule
    \end{tabular}
    \caption{Results from fine-tuning GPT3 on full and ablated versions of the prompt. Bottom sections show our baselines, Logistic Regression (LR) and vanilla GPT4/GPT3.5. All rows with (full) show models that were trained on full input prompt (Table \ref{tbl:prompt-example}). Rows with numbers, e.g. (1), etc. are ablation models trained with those parts of the prompt. Metrics are: F1, ROC-score over logits for ``yes'' tokens, Recall@10 (R@10) of each meeting (i.e. we surface the 10 most likely newsworthy items, count recall) and Mean Reciprocal Rank (MRR) of newsworthy policies, per meeting. }
    \label{tab:gpt3-results}
\end{table}

\subsection{Results and Insights}
\label{sct:sct-2:results}

In order to jointly model numerical and textual features, we choose to format our features jointly as a prompt. The structure of our full prompt is shown in Table \ref{tbl:prompt-example}, and it includes all features listed in Table \ref{tab:features}. We limit the size of the prompt by providing only the first 50 words of the text fields (besides ``proposal text''). We do not notice any impact of this truncation in early experimentation. We use this prompt to fine-tune the GPT3-Babbage model, shown to be a robust classifier \cite{spangher2023identifying}, outperforming architectures designed for text classification \cite{spangher2021multitask}. 

\paragraph{Policy text is the most predictive newsworthiness attribute, followed by meeting discussion and then public comment.}

In our first set of experiments, we ablate the prompt to explore which components of the policy are the most important for assessing newsworthiness.
We perform a temporally-based train/test split hinging on 2021/1/1. We balance our training set, with $n_{train}=641/627$ ($Y(p)=1/0$), and leave our test set unbalanced, with $n_{test}=180/2310$. We perform a time-based split rather than a randomized split because our goal is ultimately to build a model that can predict future newsworthy items. 

\begin{table}[t]
    \centering
    \begin{tabularx}{\linewidth}{lrrrrr}
        \toprule
         Train & F1 & ROC &  MRR & R@10 & n\\
         \cmidrule(lr){1-1}\cmidrule(lr){2-6}
         '13-'21 & 25.4 & 75.9 & .26 &  64.4 & 1,595 \\
         '13-'20 & 18.9 & 68.8 & .22 &  52.8 & 1,289 \\
         '13-'19 & 21.8 & 69.9 & .22 &  53.9 & 1,084 \\
         '13-'18 & 19.5 & 67.8 & .23 &  55.0 & 867 \\
         '13-'17 & 17.9 & 66.1 & .22 &  52.2 & 693 \\
         \bottomrule
    \end{tabularx}
    \caption{We alter the training split date cutoffs to be prior to Jan 1st on each of those years to test whether GPT is learning to fit to specific newsworthy events (e.g. ``COVID-19'') too well, or whether it is picking up broader newsworthy trends. For definitions of metrics, see Table \ref{tab:gpt3-results}.}
    \label{tbl:newsworthiness-by-time}
\end{table}

\begin{table}
    \centering
    \begin{tabularx}{\linewidth}{Xlr}
    \toprule
    Task & Metric & Score \\ 
    \midrule
    \multirowcell{2}[0pt][l]{Identify Newsworthy\\ \quad Policies} & Human F1 & 63.2 \\
    & \textit{(Model F1)} & \textit{(58.9)} \\
    & Cohen's $\kappa$ & 36.3\\ 
    \midrule
    \multirowcell{3}[0pt][l]{Use top $k$=10 \\ \quad as recommendation\\ \quad system} & Preference & 84\% \\ 
    & ID Accuracy & 74.2\% \\ 
    & Cohen's $\kappa$ & 60.0\\ 
    \bottomrule
    \end{tabularx}
    \caption{Results from human evaluation. Top row: journalists identify real newsworthy policies, by meeting, given a balanced dataset of 33\% $Y(p)=1$ and 66\% $Y(p)=0$ policies. Model f1-score is much higher than Table \ref{tab:gpt3-results} because this is a balanced sample. Bottom row: preference test for lists of newsworthy minutes (generated viaour models vs. random) and identification (ID) accuracy for list-origin.}
    \label{tbl:human-eval}
\end{table}

We find that the full prompt performs the best across all metrics we considered, but only marginally. As expected, ablating ``Public Comment'' from the prompt barely impacts performance, while ablating all ``meeting info.'' impacts a little more. Removing ``policy text'' information, thus forcing the model to just rely on meeting text alone impacts performance dramatically. GPT3, unsurprisingly, outperforms a very simple classifier, TFIDF+Logistic Regression (LR in Table \ref{tab:gpt3-results}), but not by much, indicating that there might be simple textual cues that we are learning. 

\paragraph{GPT4 might be capturing national newsworthiness trends.} Vanilla GPT4 outperformed our expectation. We had hypothesized that many of SFChron's newsworthiness judgements on SFBOS were local. GPT4 underperforms most other classifiers, but not by much. Manual analysis we perform finds that many errors were GPT4 failing to identify \textit{locally newsworthy} items (e.g. ``local scooter ban'', local street renaming) and that many correct predictions were made on \textit{nationally} newsworthy trends (i.e. ``COVID-19 responses''). There are two likely conclusions: (1) SFChron has major overlaps for newsworthiness judgements with national newspapers, and (2) general newsworthy language and framing is \textit{also} used for local newsworthiness.

\paragraph{Newsworthiness judgements are surprisingly consistent across time, with \textit{one major exception}.}
Table \ref{tab:most-likely-words} and Table \ref{tbl:top-topics} show that words related to specific events (e.g. those related to ``COVID-19'') are reflected in the perceived newsworthiness of policy: is the model fitting to a specific event (e.g. ``COVID-19'') that happens to be newsworthy in our training and test data, or is it learning  either (1) larger event-types (e.g. pandemics more generally,  like ``ebola'', are recurrent and newsworthy) or (2) newsworthy language patterns and other non-semantic attributes (e.g. framing)? 

To test this question, we retrain our model and increasingly restrict the date cutoffs of our training set to ask whether a model would correctly predict the newsworthiness of policies pertaining to specific events (e.g. ``COVID-19'') if the likelihood of them being in the dataset were to decrease. We show in Table \ref{tbl:newsworthiness-by-time} that, except for a dropoff after excluding data from 2021, our performance does not significantly change. 

To test whether this is the result of GPT3's pretraining, \textit{\ul{we test and are able to replicate these findings with baseline Logistic Regression models}}. An error analysis shows that \ul{``COVID-19''-related news was the least likely to be predicted correctly, and is the main contributor} to this performance decrease, whereas there are numerous other specific events that emerge (e.g. environmental, transportation-related, fire-arms related events.) that our models predict correctly. We take this as evidence that \textit{major} anomalous events, like COVID-19 specifically, do become newsworthy and are unpredictable given our current approach. \textit{\ul{This highlights an important limitation of our approach}}, as mentioned in Section \ref{sct:sct-2}. These need to be taken into account if these tools are deployed: they must be used along with other models better tuned to these blind spots. 

\paragraph{Human journalists find our newsworthiness judgements predictable and helpful.}

Finally, we recruit two expert journalists\footnote{Combined have $>40$ years of newsroom experience} and conduct human experiments with two aims: (1) is our ``newsworthiness'' definition repeatable and (2) are our models helpful? For the first, we test how well \textit{humans} able to identify newsworthy SFBOS policies. We construct a dataset by taking newsworthy policies from SFBOS meetings in our test set and a sampling nonnewsworthy policies in a 1-to-2 ratio of $Y(p)=1$, $0$. As shown in Table \ref{tbl:human-eval}, our best models achieve 58.9 F1-score on this dataset, and humans score almost equivalently. It's tempting to think our models have reached a ceiling; however, the journalists are not San Francisco-based, and  are thus untrained, compared to our models.

To test how useful these models can be, we surface $10$ policies from each meeting and ask journalists to (a) indicate which policies they might write about and (b) guess whether the list was a newsworthiness list or a random sample (they were told that it was a secondary method, not random). We found, for (a), that journalists preferred our lists to random 84\% of the time, and for (b) were able to guess which list was generated via our method 74\% of the time.

\section{Related Works}

\paragraph{Local Policy and Link Prediction} We are not the first to gather and analyze local policy discussions at scale \cite{sorens2008us, brown2021council, maxfield2022councils, barari2023localview}. Nor are we the first to study broad effects of journalism on local government \cite{hamilton2016democracy, baekgaard2014local, garz2017politicians}. Besides qualitative work \cite{felt2015incessant}, or work focused on social media \cite{graham2015role}, we believe we are one of the first to link news coverage to \textit{specific} policy. 

Link prediction is a well studied field \cite{kumar2020link}. PRMs were introduced \cite{getoor2002learning, taskar2003link} as a way of modeling attributes, but often suffered from high computational complexities. Our approach (a) uses a relatively small dataset and (b) uses entirely supervised models to ultimately make PRMs tractable here.

\paragraph{Newsworthiness prediction} has been approached in different ways. \cite{spangher2021modeling} and \cite{nishal2022crowd} sought to learn distant signals for document newsworthiness: either by classifying article layout in newspapers or by collecting attributes from crowd-workers, like ``surprising'', ``impactful''. Our work more directly addresses the question ``will this be written about?'' and allows us to study it in a data-driven manner.

This task is also called \textit{lead generation} \cite{cohen2011computational}. Of existing approaches to lead-generation, one is given by \citet{diakopoulos2010diamonds}, where a piece of content's \textit{relevance} to a given topic, its \textit{uniqueness}, and its \textit{sentiment} is quantified. Then, these metrics surface tweets related to presidential speeches. Such metrics-based systems can be interpretable, but can also miss newsworthy items that are not ranked highly by such metrics. Our work might benefit from including these metrics, and our dataset might learn to rank them well among our other features.

\section{Discussion and Conclusion}
\label{sct:discussion-and-conclusion}

In summary, we established links between a large corpus of news articles and local policy proposals we did so using a classical method, probabilistic relational modeling, that outperformed retrieval-based methods and embedding-based methods with only a small amount of annotated data. We used the assessed newsworthiness of prior articles to build models to predict the newsworthiness of articles. We found that the performance of our models did not degrade over time, and we found that expert journalists agreed with our newsworthiness assessments and found our tools helpful.

Our work faces many limitations, which we discuss in Sections \ref{sct:sct-1:results} and \ref{sct:sct-2}. 
We hope this work opens the door to future work applying these concepts, and testing them in more diverse geographic areas.

\section{Ethics Statement}

\subsection{Limitations}

We discuss a number of methodological limitations throughout our work, namely: (1) our assumptions as to linking articles give us an overreliance on lexical overlap, which is a bias our annotators might \textit{also} share based on how they chose to retrieve (article, policy) pairs. (2) Past newsworthiness might not always generalize, and might degrade more over time. There are other limitations that exist, though. The datasets we used are all in English, and local to one geography, thus are possibly not representative.

We must view our work in newsworthiness prediction with the important caveat that non-Western news outlets may not follow the patterns. We might face fundamental differences in prediction ability or problem framing if we attempt to do such work in other languages.

\subsection{Risks}

Since we constructed our datasets using well-trusted news outlets and public meetings, we assumed that every informational sentence was factual, to the best of the journalist's ability, and honestly constructed. We have no guarantees that such a newsworthiness system would work in a setting where the journalist is acting adversarially.

There is a risk that, if such a work were used in a larger news domain, it could fall prey to learning newsworthiness of misinformation or disinformation. This risk is acute in the news domain, where fake news outlets peddle false stories that attempt to \textit{look} true \cite{boyd2018characterizing, spangher2020characterizing, spangher2018analysis}. We have not experimented how our classifiers would function in such a domain. 

We used OpenAI Finetuning to train the GPT3 variants. We recognize that OpenAI is not transparent about its training process, and this might reduce the reproducibility of our process. We also recognize that it owns the models, and thus we cannot release them publicly. Both of these thrusts are anti-science and anti-openness and we disagree with them on principle. However, their models are still useful in a black-box sense for giving strong baselines for predictive problems and drawing scientific conclusions about hypotheses. By the camera ready, we will work to reduce these anti-science thrusts by experimenting with and releasing open sourced LMs. We experimented with them using DeepSpeed to run the GPT-Neo 6.7 model and the GPT Juno model on a V100 GPU. However, due to time constraints we were not able to get them working in time for submission. However, based on available evidence,\footnote{\url{https://blog.eleuther.ai/gpt3-model-sizes/}} we expect them to work at a similar capacity and will report results on them separately when we do.

\subsection{Licensing}

The San Francisco Board of Supervisors dataset we used is released without any restrictions. We have had independent lawyers at a major media company ascertain that this dataset was low risk for copyright infringement. We do not release the San Francisco Chronicle dataset that we gathered, but we do release relevant URLs, which are public domain, and scripts for accessing the Common Crawl.

\subsection{Computational Resources}

The experiments in our paper required minimal computational resources. We used a laptop computer to run baseline logistic regression and TF-IDF matching experiments. We used OpenAI's fine-tuning and prompting architecture to train GPT3 models.

\subsection{Annotators}

We recruited annotators from two major newspapers that partnered with our institution during this work. They consented to the experiment and were paid at above \$20 an hour. Both spent more than 5 years at their organizations. Neither organization is in the same locality as the San Francisco Chronicle. Both annotators are male. Both identify as cis-gender. Both are over 30 years old. This work passed a university Institutional Review Board.

\bibliography{anthology,custom}
\bibliographystyle{acl_natbib}

\appendix

\begin{table*}[t]
\centering
\begin{tabular}{llllll}
\toprule
City Lawsuits &                           Tax/Revenue &                          Basic Services &                            Environment &                        COVID-19 &                          Hearings \\
\cmidrule(lr){1-1}\cmidrule(lr){2-2}\cmidrule(lr){3-3}\cmidrule(lr){4-4}\cmidrule(lr){5-5}\cmidrule(lr){6-6}
\cellcolor{purple!44} francisco &         \cellcolor{purple!40} <number> & \cellcolor{purple!30} department &      \cellcolor{green!23} planning & \cellcolor{green!42} ordinance &      \cellcolor{green!49} health \\
       \cellcolor{purple!44} san &      \cellcolor{purple!40} exceed &      \cellcolor{purple!30} grant &          \cellcolor{green!23} code &       \cellcolor{green!42} tax &     \cellcolor{green!49} hearing \\
      \cellcolor{purple!44} city &        \cellcolor{purple!40} city &    \cellcolor{purple!30} housing &      \cellcolor{green!23} findings &      \cellcolor{green!42} tent &        \cellcolor{green!49} case \\
    \cellcolor{purple!44} county &    \cellcolor{purple!40} contract &    \cellcolor{purple!30} program & \cellcolor{green!23} environmental &        \cellcolor{green!42} hotel &  \cellcolor{green!49} commission \\
   \cellcolor{purple!44} lawsuit & \cellcolor{purple!40} authorizing &     \cellcolor{purple!30} health &        \cellcolor{green!23} street & \cellcolor{green!42} emergency &       \cellcolor{green!49} filed \\
\cellcolor{purple!44} settlement &       \cellcolor{purple!40} bonds &   \cellcolor{purple!30} services &       \cellcolor{green!23} section &     \cellcolor{green!42} covid-19 &       \cellcolor{green!49} board \\
  \cellcolor{purple!44} district &     \cellcolor{purple!40} revenue & \cellcolor{purple!30} resolution &          \cellcolor{green!23} plan &  \cellcolor{green!42} business &     \cellcolor{green!49} federal \\
     \cellcolor{purple!44} filed &    \cellcolor{purple!40} services & \cellcolor{purple!30} california &           \cellcolor{green!23} act &  \cellcolor{green!42} election & \cellcolor{green!49} supervisors \\
\bottomrule
\end{tabular}
\caption{Selection of top topics obtained by running LDA with $k=10$. Color-coding shows the likelihood of a newsworthy city council meeting minute containing a topic, with \textcolor{green}{\textbf{green}} being more likely and \textcolor{purple}{purple} being less likely. Titles are inferred topics.}
\label{tbl:top-topics}
\end{table*}

\section{Descriptive Statistics}
\label{app:descriptive-statistics}

In this appendix, we start by giving some more detailed statistics of our newsworthiness analysis. Then, we will discuss some data-processing challenges that we faced and overcame. We will discuss how we aligned transcripts with segments of the meetings, and we will discuss in more detail how we found and joined public commenters.

\subsection{More Link Analysis}

\begin{figure}[t]
    \centering
    \includegraphics[width=\linewidth]{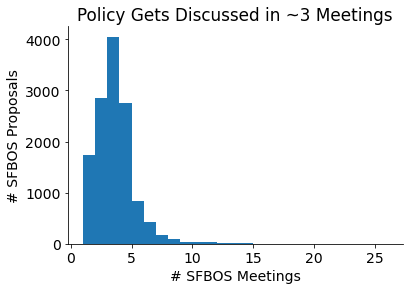}
    \caption{Amount of times policy-items get discussed in SFBOS meetings. Items go from proposals to bills and then get passed.}
    \label{fig:meetings-per-policy-item}
\end{figure}

In Figure \ref{fig:meetings-per-policy-item}, we show the number of times a policy is presented at each meeting, and find a median of 3 times. This aligns with our understanding of how policy progresses from SFBOS; it is introduced, it must be discussed and then it might pass. In cases where a policy is only discussed 1-2 times, it's more likely that it did not pass.

\begin{figure}[t]
    \centering
    \includegraphics[width=\linewidth]{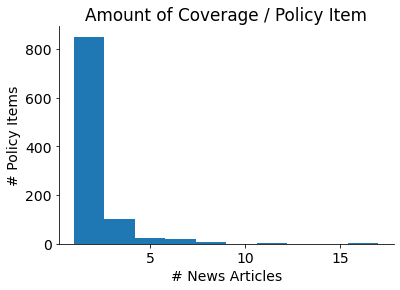}
    \caption{Number of news articles per policy item. Items get covered on average 1.8 times.}
    \label{fig:news-articles-per-policy-item}
\end{figure}

We also examine the amount of coverage given to policy items. As shown in Figure \ref{fig:news-articles-per-policy-item}, most policy items are covered between 0-1 times. However, some policy items are covered many, many times. Table \ref{tab:top-covered-bills} shows the bills that have the most coverage. We see a combination of ``COVID''-related bills, ``nominations'' and ``transportation''-related bills. While these bills do not materially affect our newsworthiness considerations, since they are more anomalies, they do provide us an opportunity to observe how coverage unfolds over time. In the future, such work could be combined with \cite{spangher2022newsedits}, \cite{spangher2023identifying} and \cite{spangher2023sequentially} to provide more of a step-by-step analysis of how coverage of especially newsworthy policies unfolds and grows over time.

\subsection{More newsworthiness Analysis}

We show in this section additional results from our newsworthiness modeling. In Table \ref{tbl:top-topics}, we performed Latent Dirichlet Allocation \cite{blei2003latent} with the number of topics set to $k=8$, as we saw a replication of topics after that. We then assigned each policy item to it's most highly-weighted topic, and counted the number of newsworthy and non-newsworthy policy items associated with each topic. We rank-order topics by the top most-newsworthy topics and the least most newsworthy topics, and shows the top 3 and bottom 3, color-coding them by their newsworthiness ranking.

As can be seen, topics like ``COVID-19'' and ``Environment'' are more present in newsworthy items, compared with ``City Lawsuits''. We assign titles to the topics based on a manual assessment.

\begin{table*}[t]
    \centering
\begin{tabularx}{\linewidth}{>{\hangindent=1.2em}Xr}
\toprule
 Top Policies by News Coverage & {\#} \\
\midrule
 Commending Supervisor London Breed. Resolution Commending and Honoring Supervisor London Breed for her distinguished service as a Supervisor of the City and County of San Francisco. & 17 \\
Transportation, Public Works Codes - Unauthorized Powered Scooter Violations, Powered Scooter Share Program. Ordinance amending Division I of the Transportation Code to establish a violation for Powered Scooters that are a part of a Powered Scooter Share Program, to be parked, left standing, or left... & 16 \\
Emergency Ordinance - Limiting COVID-19 Impacts through Safe Shelter Options. Emergency ordinance to require the City to secure 8,250 private rooms by April 26, 2020, through service agreements with hotels and motels for use as temporary quarantine facilities for people currently experiencing homele... & 16 \\
Nomination Process and Appointment of a Successor Mayor. Motion to take nominations and appoint a successor Mayor to fill a vacancy in the Office of the Mayor, during a Committee of the Whole hearing of the Board of Supervisors of the City and County of San Francisco on January 23, 2018. & 14 \\
 Mario Woods Remembrance Day - July 22. Resolution declaring July 22 as Mario Woods Remembrance Day in the City and County of San Francisco. & 12 \\
Approving Submission of Sales Tax to Support Caltrain Service - November 3, 2020, Election. Resolution approving the Peninsula Corridor Joint Powers Board's placement of a three-county measure to impose a one-eighth of one percent (0.125\%) retail transactions and use tax to be used for operating and... & 12 \\
Park Code - Golden Gate Park Access and Safety Program - Slow Street Road Closures - Modified Configuration. Ordinance amending the Park Code to adopt the Golden Gate Park Access and Safety Program, which includes restricting private vehicles on certain slow street segments in Golden Gate Park inclu... & 11 \\
\bottomrule
\end{tabularx}
    \caption{Top SFBOS policies, by the number of times they were covered in the SFChron (\#). Includes a mix of office-related, transportation bills and COVID bills.}
    \label{tab:top-covered-bills}
\end{table*}

\section{Aligning meeting transcripts with video}
\label{app:aligning-meeting-transcripts}

\begin{figure*}[t]
    \centering
    \includegraphics[width=\textwidth]{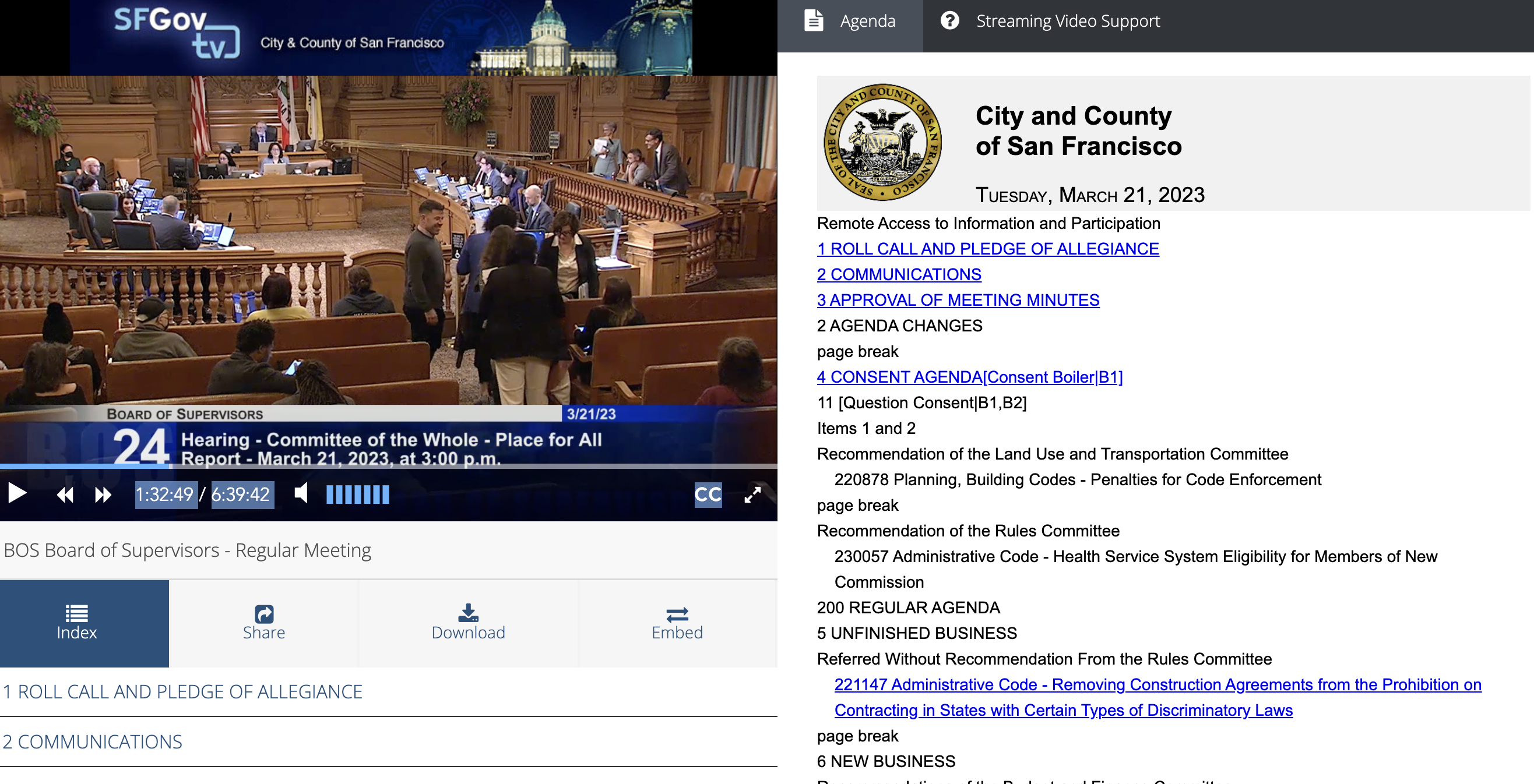}
    \caption{A screen shot of the SFBOS video-hosting website for the URL: \url{https://sanfrancisco.granicus.com/player/clip/43243}. Seen on the left is the video, from which we are able to download an audio \texttt{.mp3} file. On the right is an agenda items from which we can parse timestamps for policy discussions.}
    \label{fig:granicus-screenshot}
\end{figure*}

We collect $3,200$ hours of video data for SFBOS meetings from their hosted service\footnote{Called \texttt{Granicus}, which is a service provider used for many local governments. } In this section, we describe how we parse the sections of the video that correspond to the policy-items. 


Figure \ref{fig:granicus-screenshot} shows an example landing page for one SFBOS meeting, held on March 21, 2023. As seen on the left-hand side, the video is shown. On the right-hand side, a nested, hyperlinked agenda document is shown. Capital-letter headers are canonical meeting sections, and are relatively constant across meetings. 

\begin{table}[t]
    \centering
    \begin{tabularx}{\linewidth}{>{\hangindent=2em}X}
    \toprule
    Transitional Phrase\\
    \midrule
     Madam clerk, please call item next item item 33.\\
     Without objection, this resolution is adopted unanimously. Item number nine. Item nine.\\
     Those items are adopted unanimously. Next item, please.\\
     Item number 61. Item 61.\\
     Without objection the resolution is adopted unanimously. Item 44. Item 44\\
     Next item. Item 24.\\
     Without objection, the resolution is adopted unanimously. Next item. Item 52.\\
     Madam clerk, please call item the following items together item 44 45 and 50 and item 54.\\
     Madam Clerk, would you call item 5 please?\\
     Without objection, this resolution is adopted unanimously. Madam Clerk, item number 18.\\
     Madam Clerk, please call the next item.\\
     \bottomrule
     \end{tabularx}
     \caption{Examples of agenda-item transitions that we identified and then used to parse the agenda..}
     \label{tab:transition-examples}
\end{table}

\begin{table*}[t]
\centering 
    \begin{tabularx}{\linewidth}{>{\hsize=.9\hsize \hangindent=2em}X>{\hsize=1.1\hsize \hangindent=2em}X}
\toprule
Policy & Public Comment Speaker (transcribed text) \\
\midrule
130049 Resolution supporting Senator Dianne Feinstein’s Assault Weapons Regulatory Act of 2013. & Good day, Supervisors. My name is \censor{Robert Green}. I'd like to start by saying that I do not own any firearms, and I do not oppose sensible gun control legislation. Yet I rise today in opposition to Item... \\
130151 Resolution opposing the indefinite detention provisions of the National Defense Authorization Act, instructing public agencies to decline requests by Federal agencies acting under detention pow... & Hi, I'm \censor{Aubrey Friedman} from the Libertarian Party of San Francisco and we fully support David Chu's resolution against the detention provisions of the NDAA. Under the guise of the War on Terror, The... \\
130257 Resolution standing with Muslim and Arab communities in the face of anti-Arab and anti-Muslim bus advertisements. & Hello, my name is \censor{Christina Sinha} and I'm a staff attorney at the Asian Law Caucus. And I'd like to speak today about the racist advertisements, the anti-Muslim and anti-Arab advertisements that have ... \\
130425 Resolution authorizing the Department of Public Library to retroactively accept and expend a grant in the amount of up to \$750,000 of in-kind gifts, services, and cash monies from the Friends o... & Good afternoon, Supervisors. Stop the corporate rape of the public library. Don't give money to the Friends of the Library. Don't accept money from the Friends of the Library. Before we begin, we shou... \\
131071 Accept and Expend Grant - Library Programs - Friends of Public Library - Up to \$720,000 - FY2013-2014 & Good afternoon. I'm \censor{Peter Warfield}, Executive Director of Library Users Association. I would like to ask the supervisors to have a hearing on library plans and priorities and performance. The library,... \\
\bottomrule
\end{tabularx}
\caption{Sampling of public comments, mapped to the policies we infer that they are supporting.}
\label{tbl:public-comment-samples}
\end{table*}

Some of the lines are shaded in blue, meaning that they are hyperlinked to a timestamp in the video. In the agenda, any line that starts with a 6-digit code refers to discussions around a policy-item that the SFBOS wishes to pass. Some of the policy-item lines have hyperlinks pertaining to them while others do not.

\begin{table*}[t]
\centering
\begin{tabularx}{\linewidth}{>{\hangindent=2em}Xrll}
\toprule
Meeting Section &  Time (Min) & \# Policies & \# Speakers\\
\midrule
COMMITTEE REPORT                         &     38.1 &      3 (+/- 5)  & 5 (+/- 9)\\
SPECIAL ORDER                            &     29.3 &      5 (+/- 8)  & 15 (+/- 22)\\
PUBLIC COMMENT                           &     23.9 &      0 (+/- 1)  & 16 (+/- 13)\\
CONSENT AGENDA                           &      4.6 &      3 (+/- 4)  & 3 (+/- 5) \\
NEW BUSINESS                             &      3.5 &      11 (+/- 9) & 10 (+/- 10)\\
REGULAR AGENDA                           &      2.5 &      5 (+/- 7)  & 4 (+/- 5) \\
IMPERATIVE AGENDA                        &      1.8 &      2 (+/- 3)  & 3 (+/- 7) \\
FOR ADOPTION WITHOUT COMMITTEE REFERENCE &      1.4 &      5 (+/- 4)  & 7 (+/- 9)\\
UNFINISHED BUSINESS                      &      1.3 &      3 (+/- 6)  & 3 (+/- 4) \\
ROLL CALL                                &      1.0 &      2 (+/- 3)  & 13 (+/- 14) \\
PROPOSED RESOLUTION                      &      0.4 &      2 (+/- 3)  & 1 (+/- 3) \\
COMMUNICATION                            &      0.3 &      0 (+/- 1)  & 3 (+/- 4) \\
APPROVAL OF MEETING MINUTE               &      0.3 &      0 (+/- 1)  & 1 (+/- 1) \\
AGENDA CHANGE                            &      0.2 &      0 (+/- 1)  & 2 (+/- 2) \\
PROPOSED ORDINANCE                       &      0.2 &      1 (+/- 2)  & 2 (+/- 12) \\
ADJOURNMENT                              &      0.0 &      1 (+/- 2)  & 2 (+/- 2)\\
\bottomrule
\end{tabularx}
\caption{Top-level parts of SFBOS meetings and the average amount of time spent on each one, according to our inferred timestamps. Also shown are the mean \# of policy items discussed in each part, on average, as indicated by the agenda, and the mean \# of speakers per section, as per diarization.}
\label{tbl:header-timestamps}
\end{table*}

We manually examined agendas. Many links were missing simply because several policies were discussed together in the agenda. However, others seemed to be missing randomly, leading us to believe that the agenda hyperlinks were incompletely linked. 

We wish to reconstruct as completely as we can the time-stamped agenda so that we could get an accurate segmentation for the meeting, so we aim to fill in the missing agenda items. We explore a very simple hypothesis: we assume that meetings were highly organized, and there were consistent phrases used to transition to different agenda items. 

So, we seek to classify transitionary phrases. We train a classifier that takes as input a list of diarized transcriptions, $t$, which each have their own start and end times $(t_s, t_e)$ annotated from the transcription process, and predicts:

\begin{equation}
Y(t) =
\begin{cases}
  1, & \text{ if } \exists \text{ hyperlink } \in (t_s, t_e) \\
  0, & \text{otherwise}
\end{cases}
\label{eq:newsworthiness-label}
\end{equation}

We recognize this is noisy, as $Y(t)$ can $=0$ if both: (A) a segment is not a transition \textit{or} (B) it is simply missing a transition. However, we train a simple classifier using bag-of-words representations for diarized segments and logistic regression, and we achieve f1-score=.86 on held-out data. We analyze the outputs of the classifier and see that it discovers relevant transitional speech, see Table \ref{tab:transition-examples} for examples. 

Having labeled our transcripts with each diarized segment's likelihood of being transitional speech, we then iterate through each transcript and peg each unlabeled block to either: (a) the next most likely transitional segment or (b) the previous item's segment, if no segment exists above $.8$ likelihood. In this way, we allow multiple agenda items to be discussed in the same short segment.

This is an exceedingly simple approach that does not consider semantic similarities between the meeting agenda as an approach like, say, dynamic time warping \cite{} might. We are confident that our approach could be improved, and maybe in future work improvements could result in additional signal being observed.

\section{Additional Joining Information}
\label{sct:missing-policies}

When we extract agendas in the SFBOS video viewer, as shown on the righthand side of Figure \ref{fig:granicus-screenshot}, we find that we are able to retrieve a total of only 10,877 out of 13,089 policies which were listed on the SFBOS legislative calendar website as being discussed during meetings. This is strictly a subset. All policies gathered from video viewer agendas are listed in the SFBOS legislative calendar website.

It's likely that this discrepancy results from policies that were introduced but did not make it past preliminary stages of investigation. For instance, here is an example of a proposal that was listed in the legislative calendar website: \url{https://sfgov.legistar.com/LegislationDetail.aspx?ID=2070276&GUID=D31163A0-D5F8-41E7-AB90-4DECAF9E6693} as having been presented during a SFBOS meeting on 11/18/2014. However, the actions associated with that item, as told in the website, are: ``RECEIVED AND ASSIGNED'', ``REFERRED TO DEPARTMENT'', ``TRANSFERRED'', and ``FILED PURSUANT TO RULE 3.41''. Here is the video page of the 11/18/2014 meeting: \url{https://sanfrancisco.granicus.com/player/clip/21460}. As can be seen, policy number 141197 is not listed in the agenda. That is different from, say, this proposal: \url{https://sfgov.legistar.com/LegislationDetail.aspx?ID=6122328&GUID=B5231DEE-0596-463F-8934-A84468D131ED}, which was ``CONTINUED'' and ``HEARD AND FILED'', with meeting details associated with each one. 

So, a logical explanation is these policies that were never brought to discussion during meetings, thus they do not appear in meeting agendas. However, we cannot discount the possibility that errors were made in creating the agendas. In this case, we were not able to track policies that were genuinely discussed. Nevertheless, we will refer to these policies as ``unpassed-policies''.

This affects 164 unpassed-policies that we have identified as being covered by SFChron, out of a total of 1,015 policies, or 16\% of newsworthy policies. These unpassed-policies were covered 298 times, out of a total of 2001 articles. We give a sampling of these missing newsworthy unpassed-policies in Table \ref{tbl:newsworthy-missing-agenda-items}. While it's entirely likely that the \textit{fact} that these policies were not discussed \textit{lead} to them being newsworthy, we do not consider such a distinction in our modeling. We leave this to future work.

\begin{table*}[t]
\begin{tabularx}{\linewidth}{>{\hangindent=2em}X}
\toprule
Sample of Newsworthy Policies that were not found in SFBOS Video Agendas \\
\midrule
Committee of the Whole - Standing Briefings Related to the COVID-19 Health Emergency Response on Board Tuesdays at 3:00 p.m.. Motion directing the Clerk of the Board of Supervisors to schedule standing Committee of the Whole hearings every Tuesday that the Board of Supervisors has a regular meeting ... \\
Hearing - Federal Budget Cuts to Health Care, Immigration Services, Homeless Services, and Services for the LGBTQ Community. Hearing on the federal budget cuts to health care, immigration services for undocumented San Franciscans, services for the LGBTQ community, homeless services, and cuts to serv... \\
 Concurring in the Continuation of the Declaration of a Local Health Emergency - Monkeypox Virus Outbreak. Motion concurring in the continuance of the San Francisco Health Officer's August 1, 2022, Declaration of Local Health Emergency regarding the outbreak of the Monkeypox virus. \\
Appropriation - Department of Building Inspection Fund to Department of Emergency Management for Tall Building Seismic Safety Project - \$250,000. Ordinance appropriating \$250,000 of fund balance in the Department of Building Inspection fund to Department of Emergency Management for Tall Building Sei... \\
Hearing - 2022 Aging and Disability Affordable Housing Needs Assessment Report. Hearing requesting the key findings and recommendations made in the 2022 Aging and Disability Affordable Housing Needs Assessment Report; and requesting the Department of Disability and Aging Service, Mayor's Office on H... \\
\bottomrule
\end{tabularx}
\caption{Sample of newsworthy policies that were not found in agendas listed on SFBOS video page viewers. We believe that the most were never discussed, but there could be errors in creating the agenda.}
\label{tbl:newsworthy-missing-agenda-items}
\end{table*}

\section{Additional Meeting Exploration}
\label{app:additional-meeting-eda}

Having parsed each agenda and time-pegged each line-item in the agenda, we are able to roll up the time spent in each section-header. Table \ref{tbl:header-timestamps} shows the length of time spent in each section.

As can be seen, the ``PUBLIC COMMENT'' section occupies a major part of meetings, in terms of the amount of time spent in each meeting, and yet very few policy-proposals are explicitly discussed during this period. We hypothesize that public comment is a potentially newsworthy period in the meeting, where members of the public are able to raise the emotional tenor of a piece of policy (which makes for good news-writing \cite{uribe2007aresensational}). So, we attempt to join publicly-made comments to entire text of the policy discussion. 

As discussed in the main body, we defined ``public commenters'' as members of the public who only speak during the ``PUBLIC COMMENT'' section of the meeting. Given timestamps for this section, and speaker diarization, we are able to filter out all speakers besides those that speak during public comment. Next, we use word-overlap between the speaker's speech and the policy text to determine whether the speaker is addressing a particular topic. For the sake of brevity, we assume that each speaker only addresses one comment.

We show in Table \ref{public-comment-samples} some examples of public commenters. As can be seen, they address policy with a personal tenor. However, there are also comments that are rather noisy (e.g. meandering, not on topic, not very focused.) We feel that more work is needed to make the public comment section a usable part of this analysis.

\end{document}